\documentclass[runningheads]{llncs}
\usepackage{graphicx}
\usepackage{multirow}
\usepackage{subcaption} 
\usepackage{url}
\usepackage{amsmath}
\usepackage{longtable}
\usepackage{makecell}

\begin{document}
\title{Developmental Predictive Coding Model for Early Infancy Mono- and Bilingual Vocal Continual Learning}
\titlerunning{Developmental Predictive Model for Early Infancy Vocal Learning}
%
\author{Xiaodan Chen\inst{1,3}\orcidID{0009-0007-3949-5353} \and
Alexandre Pitti \inst{1,3}\orcidID{0000-0002-6541-578X} \and
Mathias Quoy\inst{1,3}\orcidID{0000-0003-3345-380X} \and
Nancy F. Chen\inst{2,3}\orcidID{0000-0003-0872-5877}}
\authorrunning{X. Chen et al.}
%
\institute{ETIS, CY Cergy-Paris Université - ENSEA – CNRS, UMR 8051, 2 Av. Adolphe Chauvin, 95300 Pontoise \\
\email{firstname.lastname@cyu.fr}\\ 
\and
A*Star, 1 Fusionopolis Way, \#20-10 Connexis North Tower, Singapore 138632
\email{nfychen@i2r.a-star.edu.sg}\\
\and
IPAL (International Research Laboratory on Artificial Intelligence), CNRS, Singapore}


\maketitle              
\begin{abstract}

Understanding how infants perceive speech sounds and language structures is still an open problem. Previous research in artificial neural networks has mainly focused on large dataset-dependent generative models, aiming to replicate language-related phenomena such as "perceptual narrowing". In this paper, we propose a novel approach using a small-sized generative neural network equipped with a continual learning mechanism based on predictive coding for mono- and bilingual speech sound learning (referred to as language sound acquisition during "critical period") and a compositional optimization mechanism for generation where no learning is involved (later infancy sound imitation). Our model prioritizes interpretability and demonstrates the advantages of online learning: Unlike deep networks requiring substantial offline training, our model continuously updates with new data, making it adaptable and responsive to changing inputs. Through experiments, we demonstrate that if second language acquisition occurs during later infancy, the challenges associated with learning a foreign language after the critical period amplify, replicating the perceptual narrowing effect.

\end{abstract}
\section{Introduction}
The discourse surrounding the origin of human speech spans across an extensive historical timeline. \cite{Rizzolatti1998} suggests that the mimetic capacity inherent in brain areas like F5 and Broca's area played a crucial role in the transition from gestural communication to speech \cite{Stoke_1981}. As such, vocalization became more than just emotional reinforcement for facial expressions \cite{Armstrong_Stokoe_Wilcox_1995}. Instead, sounds acquired a descriptive value and needed to remain consistent in identical situations. This required not only the ability to produce specific sounds but also the ability to imitate sounds emitted by others, which likely contributed to the emergence of human Broca's area from a precursor with mirror properties. This area, associated with speech production and language processing, became crucial for the control and coordination of vocalization, especially in terms of imitation and precise execution of sounds \cite{Rizzolatti1998}.

On the other hand, studies \cite{Fowler1986} suggest that listeners can quickly mimic speech, demonstrating faster reactions to syllables or gestures that align with those of the speaker. This tendency to imitate may originate from early infancy, where it likely serves a crucial developmental role. Infants must decipher the phonetic nuances of speech signals from adults to learn how to map their own motor commands
for speech reproduction. In humans, vocal imitation is a pivotal behavior that lays the foundation for language development. Studies have noted the mimicry of sounds emerge as early as 2 months of age \cite{Oller1999}. This early vocal imitation has been found to contribute to later lexical development, and more importantly, to the acquisition of second language (L2), reflecting on the later ability for reproduction of foreign speech sounds \cite{Reiterer2013}. As such, these speech sound imitation variations among individuals can lead to noticeable foreign accent disparities among late L2 learners \cite{Nguyen2015}.

However, individual variances of capacity in L1 learning is not the only influence on L2 learning. A phenomenon called perceptual narrowing witnessed across various domains within the first year of age may account for that. The term perceptual narrowing \cite{Werker1984} initially served as a descriptive phrase highlighting findings that infants exhibit greater sensitivity to various social signals early in life, including non-native speech sounds, and this sensitivity decreases over time as they become more attuned to their surrounding environments \cite{Knudsen2004}. Though lost discrimination can be regained outside the critical period with extensive and substantial training, indicating perceptual narrowing involves attenuation or reorganization rather than complete loss, some phonetic distinctions continue to be challenging compared to face discrimination even with significant training \cite{Maurer2014}. 

Such a phenomena is closely related to two key factors. 
The first one is that "minimal input"\cite{Maurer2014}, namely early foreign-language exposure\cite{Kuhl2003} is needed to maintain these initial sensitivities. A famous example is Japanese children's deficit in distinguishing difference between English -r and -l as the absence to the sounds during their early infancy due to the linguistic environment of their parents \cite{Kuhl2010}. The second one is 'critical period', a distinct subset within the concept of 'sensitive periods' \cite{Knudsen2004} that refers to a specific time window during which experience is essential for learning to take place, and the knowledge acquired during this period leads to lasting effects that are difficult to reverse \cite{Kuhl2005}. After this period, the system will not be open to, or at least will be more resistant to, reorganization or retuning at a older age. \cite{Vouloumanos2014} shows that adults' perception of speech sounds becomes constrained compared to that of newborns, as influenced by the phonetic patterns of their native language, which appears by the end of the first year after birth.

As a result, we 
formulate
the hypothesis that early infancy imitation lays foundation for later development not only of words or language but also of L2 acquisition. Based on that, we propose a simple neural network consisting of a categorization encoder and a generation decoder that learns to mimic input sounds. Nevertheless, instead of exploring the causality that better capacity of L1 learning leads to the better L2 learning, we proposed two modes of generation that corresponds to early and later infancy imitation. 
The first mode is predictive coding based on continual learning (CL). Predictive coding is based on maximizing prediction similarity through recurrent interactions\cite{Kilner2007}. Discrepancies between predicted and actual sensory input (prediction error) is used to adjust the model or prediction \cite{FRISTON20031325}. There are feedback loops where prediction errors are propagated through the system to refine predictions \cite{Friston2008}. Continual learning is a fundamental aspect of predictive coding, allowing the system to adapt and improve its predictions over time based on new sensory input \cite{Clark2013}.
The second mode is compositional optimization (CO), where we propose a hypothesis that the later childhood sound imitation depends greatly on their early infancy 'minimal input' and that the generation of later-heard sound comes from the compositionality of the sound learned during 'critical' time window. We are able to show through experiment that if the second language acquisition happens during later childhood or adulthood, the difficulty of learning foreign language after the ‘critical’ period appears, similar to perceptual narrowing.

\section{Related works}
\subsection{Perception and production reinforce imitation}
In Kuhl et al. \cite{Kuhl2000}, the role of imitation in vocal learning is emphasized as crucial for establishing the early connection between perception and production. In line with this, studies have shown that motor and sensorimotor systems can impact speech perception, even in infants who are too young to produce speech-like vocalizations \cite{Choi2023}. \cite{Imada2006} examine studies revealing that sensorimotor information regarding speech emerges prior to the developmental onset of speech-like vocalizations, as inferior frontal cortex (IFC) including Broca’s area, exhibits activation during auditory-only speech perception even in very young infants who have not yet initiated speech-like vocalizations. Concurrently, Zhao et al. in \cite{Zhao2022} illustrate that activation patterns of the IFC exhibit correlations with speech sound perception in infancy: similar activation patterns are observed when neonates and early infants perceive both native and non-native speech sounds, with a decrease in activation and perception of non-native speech sounds occurring by the end of infancy. 

\subsection{Speech sound learning neural networks}
In the following part, we discuss some neural networks for speech sound learning.

\subsubsection{Perceptual narrowing}
Schatz et al. \cite{Schatz2021} came up with a Gaussian mixture-based model that is able to reproduce perceptual narrowing. They demonstrated that merely 1 hour of input is sufficient for the emergence of distinct characteristics between Japanese and English models, with this gap amplifying with exposure to the native language. While the observation of intensified cross-linguistic contrast with increased data aligns with empirical findings in infants, it remains challenging to determine whether this correlation is attributable to dataset augmentation leading to overfitting or enhanced learning of linguistic features, as they did not involve L2 learning for further comparison and analysis based on the L1 learning model.

\subsubsection{Sensori-motor learning}
Georges et al. \cite{Georges2022} introduced a computational model of speech production featuring forward and inverse models that can be jointly trained. It encompasses several components, such as a pre-trained neural articulatory synthesizer responsible for converting articulatory parameters into speech stimuli, a DNN-based internal forward model for predicting sensory outcomes, and an internal recurrent neural network designed to recover articulatory commands from acoustic speech input. While this model has shown the capability of reproducing speech sounds trained in a self-supervised manner using raw acoustic-only speech data, limited exploration has been conducted on linguistic-related features.

\subsubsection{Other sound learning models}
The Zero Resource Challenge (ZRC) \cite{Dunbar2022} introduced the concept of replacing textual language databases with raw audio databases, drawing inspiration from the fact that infants acquire speech skills long before they can read or write. One of the four tasks proposed in the challenge, closely related to this paper, is acoustic unit discovery. Similarly, they argue that the units discovered should serve a fundamental linguistic function: distinguishing linguistic contrasts. However, most neural networks in this task employ large-scale architectures, with few exploring bilingualism.

\section{Methods}
In this section, we will introduce the encoder-decoder architecture of our generative network along with the schema of the proposed models. 

\subsection{Network architecture}

\begin{figure}
\includegraphics[width=\textwidth]{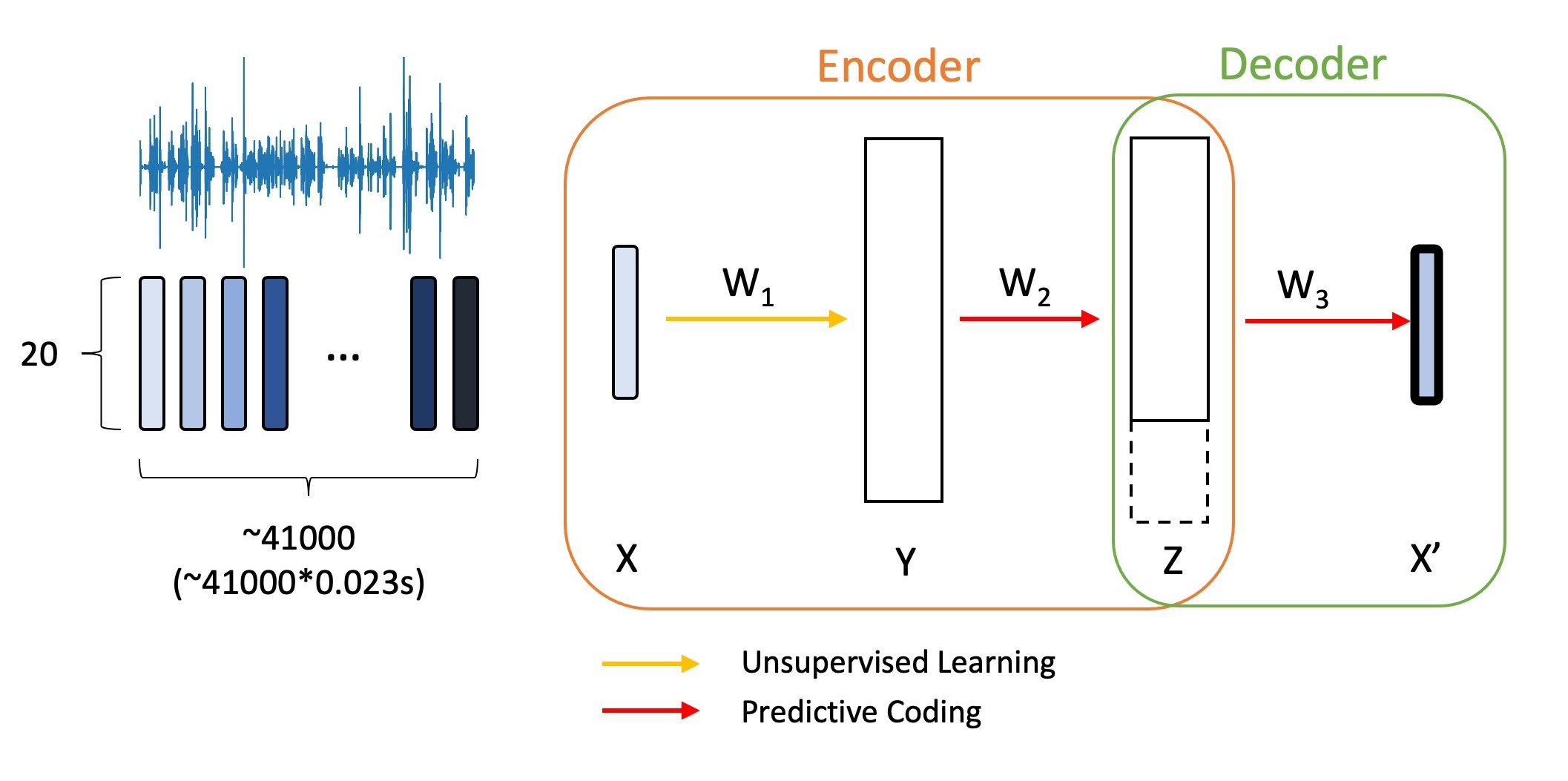}
\caption{Our proposed encoder-decoder network architecture. The yellow arrow denotes unsupervised learning whereas the red one represents predictive coding \cite{FRISTON20031325}\cite{Pitti2020}.}
\label{architecture}
\end{figure}

The network architecture, depicted in Fig. \ref{architecture}, comprises an encoder and a decoder. The input to the neural network is a 20-dimensional MFCC (Mel-frequency cepstral coefficients) vector, corresponding to a 23ms sound segment. Layer Y is a one-dimensional self-organizing map (SOM) \cite{Kohonen1982} consisting of a fixed number of 2000 neurons. The weight update rule is described in Equation \ref{equation_w1} and the output of SOM layer Y in Equation \ref{equation_y}, both adhere to the standard SOM formulation \cite{Kohonen2001}. Please note that in all equations in the following sections, lowercase letters correspond to vectors, and uppercase letters correspond to matrices.

\begin{equation}
    w_1^{ij} (t+1)=w_1^{ij} (t)+\theta(i_0,i,t) \cdot \alpha(t) \cdot (x_j- w_1^{ij} (t)) 
\label{equation_w1}
\end{equation}
where $i$ and $j$ denote the index of the neuron in Y and the index of the input, respectively. The smallest of the Euclidean distances $\| x_j - w_1^{ij} \|_{2}^{2}$ can be performed to define the best-matching unit (BMU), signified by the subscript $i_0$. The neighborhood kernel $\theta(i_0,i,t)$ known as "Mexican hat function" \cite{Gustafsson1997} or Gaussian function \cite{Kohonen2001}, involves excitatory and inhibitory lateral feedback connections, where the width of the kernel decreases over time. Similarly, the learning rate function $\alpha(t)$ is also a temporal function that decreases along training iterations \cite{Kohonen2001}. These temporal functions guarantee smoothly transition from a broad neighborhood influence at the beginning of the training to a narrower influence as training progresses. This gradual reduction helps the map adapt quickly in the initial stages and fine-tune itself in the later stages, leading to a more accurate and stable representation \cite{Kohonen1998}.

\begin{equation}
Y = \frac{1}{1 + \|X - W_{1} \|_{2}^{2}}
\label{equation_y}
\end{equation}

In contrast to layer Y, the size of layer Z is dynamic and increases when the variance of neuron activity in layer Y falls below a certain threshold value. The layer and the number of neurons required or activated in each layer is specified in Table \ref{table_layer_neuron}. The network's structure with three layers and moderate neuron counts fits well within the typical bounds for small-sized neural networks. Further details regarding the remaining components of the neural network will be provided in subsequent sections.

\begin{table}[h!]
\centering
\caption{Layers and Number of Neurons Required for the Neural Network.}
\label{table_layer_neuron}
\resizebox{0.8\textwidth}{!}{
\begin{tabular}{|c|c|c|c|c|c|}
\hline
\multirow{2}{*}{\textbf{Layer}} & \multirow{2}{*}{\textbf{Type}} & \multirow{2}{*}{\textbf{Train}} & \multicolumn{3}{c|}{\textbf{Test Periods/Languages}} \\ \cline{4-6} 
 &  &  & \textbf{English} & \textbf{French} & \textbf{Chinese} \\ \hline
\textbf{X} & MFCC input number & \makecell{$\sim$41000 \\ ($\sim$16min)} & \makecell{$\sim$20000 \\ ($\sim$8min)} & \makecell{$\sim$20000 \\ ($\sim$8min)} & \makecell{$\sim$20000 \\ ($\sim$8min)} \\ \hline
\textbf{Y} & SOM winner neuron number & 1603 & 1349 & 901 & 691 \\ \hline
\textbf{Z} & Neuron number & 1815 & 1733 & 946 & 836 \\ \hline
\end{tabular}
}
\end{table}

\subsection{Proposed reconstruction modes}

\begin{figure}[h!]
\includegraphics[width=\textwidth]{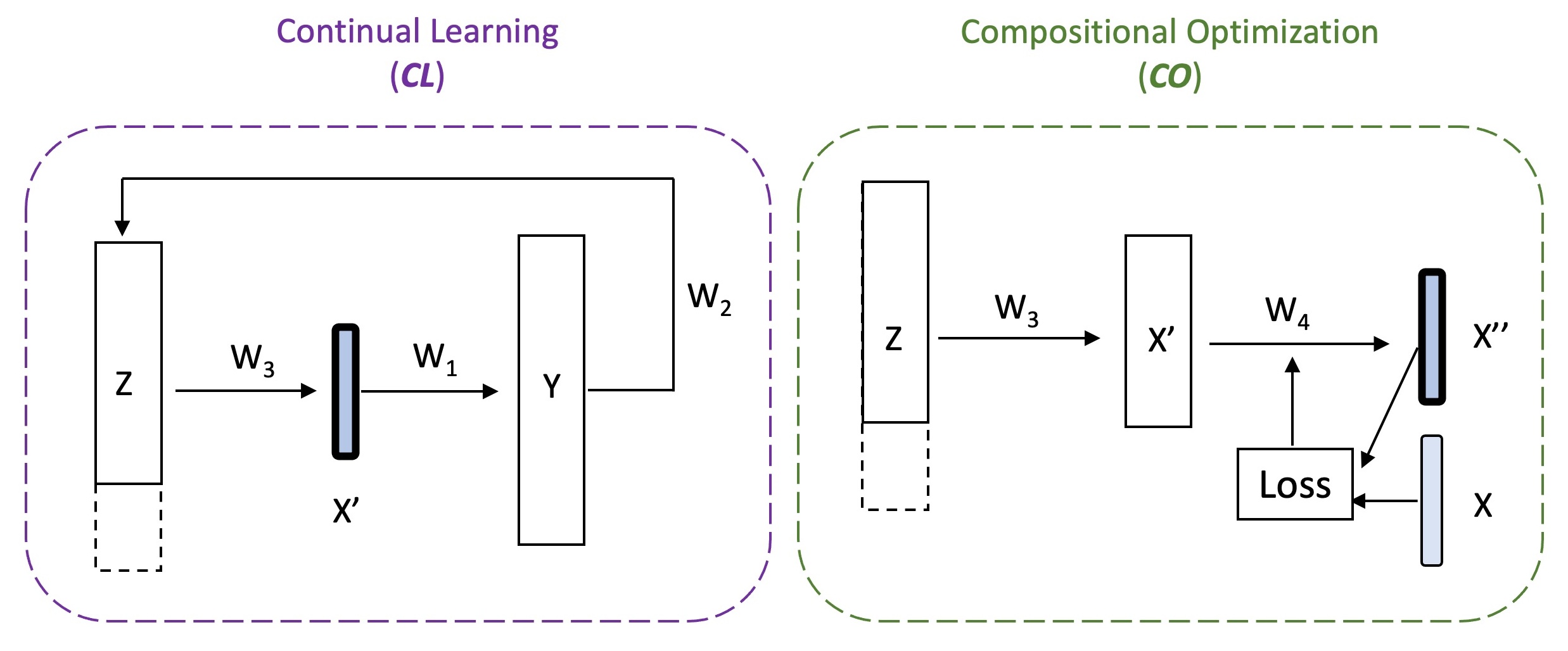}
\caption{Schema of the proposed modes. Left: Continual learning (CL) mode that simulates early infancy sound imitation, where the model generates sounds heard in a reinforced way. The SOM layer $Y$ serves as a predictor and the reconstructed output is adjusted based on sensory input. Continual learning is implemented by updating the SOM's internal representation (Equation \ref{equation_y_x'}) based on new input data $X'$ over time, allowing the system to adapt its predictions to changes in the environment and improve prediction accuracy through continual learning. Such mode exhibits a simplified hierarchical structure within predictive coding, where $Z$ adapts similarity (Equation \ref{equation_Z}) to refine predictions. Right: proposed compositional optimization (CO) mode simulating later infancy sound generation, modelizing our hypothesis that the ability to imitate sounds in later childhood is mainly influenced by the minimal input received during early infancy. Later-learned sounds are generated based on the compositional nature of sounds acquired during critical developmental periods.} 
\label{schema}
\end{figure}

According to the suggested hypothesis, the ability of children to imitate sounds later in childhood is significantly influenced by the minimal auditory input they receive during early infancy. Here, we propose a predictive coding based continual learning (CL) mode shown on the right side of Fig. \ref{schema} that captures this kind of learning.  Additionally, we assume that the sounds they later reproduce are constructed based on the composite elements of sounds (constituents) learned during a critical developmental window in early infancy. We term this "no learning" sound imitation or generation "compositionality optimization" (CO) and we propose the compositional optimization model shown on the right side of Fig. \ref{schema} to demonstrate it.

\subsubsection{Continual learning mode}
The process of predictive coding based continual learning occurs subsequently to the training of layer $Y$. 
Predictive coding involves a hierarchical structure, where higher levels generate predictions for lower levels, and prediction similarities are propagated back up the hierarchy to refine those predictions. 
Here, the self-organised map becomes a static mapper or a priori structure as top layer, used to transform causes (input) to some internal representation \cite{Friston2008}.

Consequently, the weights $W_{2}$ between layers $Y$ and $Z$ represent the pattern (internal representation) transformed from the heard sound (cause)
as shown in Equation \ref{equation_w2}, modulating the activities of $Y$, which in turn represents the pattern of the produced sound (predictions about the states of lower layers),
as shown in Equation \ref{equation_y_x'}. 

\begin{equation}
W_{2} = Y
\label{equation_w2}
\end{equation}
\begin{equation}
Y = \frac{1}{1 + \|X' - W_{1} \|_{2}^{2}}
\label{equation_y_x'}
\end{equation}

The similarity maximization in predictive coding is realised through multiplication as shown in Equation \ref{equation_Z}, 
where only the winning neuron in $Z$ becomes activated, facilitating the continual generation of new sounds to approximate the heard sound. 

\begin{equation}
Z = W_2 \cdot Y
\label{equation_Z}
\end{equation}

The equation governing the update rule for the weights $W_{3}$ connecting layer $Z$ to the output layer is detailed in Equation \ref{equation_w3}. Here, $i_0$ indicates that only the weight associated with the winning neuron in layer $Z$ and the output layer is reinforced. This reinforcement is realized in the form of adding a randomly generated Gaussian-type input to itself if this Gaussian-type input amplifies the activity of the winning neuron within layer $Z$, minimizing prediction error.

\begin{equation}
    w_{3}^{i_0} (t+1) = w_{3}^{i_0} (t) + I_{G}\qquad where\quad I_{G} \sim \mathcal{N}(0, 0.01)
\label{equation_w3}
\end{equation}

As a result, $X'$ represents the newly generated input for the next iteration, namely the reconstructed output of the current iteration, which is adjusted based on the input. In the hierarchical predictive coding structure, $X'$ is considered as the lower layer, which is then looped back to Layer $Y$, serving as a prediction that is updated to minimize the discrepancy with the actual input.

\begin{equation}
X' = \delta_{i_0}(Z) \cdot W_{3} \qquad where\quad \delta_{i_0}(Z) = \begin{cases}
    1, & \text{if } i=i_0 \\
    0, & \text{otherwise}\end{cases}
\label{equation_X'_CL}
\end{equation}

In this setup, the SOM layer $Y$ functions as a predictor, utilizing its learned mapping to generate predictions. Meanwhile, the reconstructed output $X'$ undergoes adjustments contingent upon sensory input. This process mirrors the principles of predictive coding, where continual learning is facilitated by iteratively updating the SOM's internal representation with fresh input data $X'$. This adaptability empowers the system to evolve its predictions in response to environmental changes, thus enhancing prediction accuracy over time. Through this continual learning mechanism, the model exhibits a simplified hierarchical structure within the predictive coding framework. By leveraging similarity to fine-tune predictions, it refines its predictive capabilities and adapts to dynamic environmental cues.

\subsubsection{Compositional optimization mode}
The CO mode shown on the right side of Fig. \ref{schema} utilizes the combination coefficient $W_{4}$, which is optimized through backpropagation as described in Equation \ref{equation_w4}, without involving any learning.

\begin{equation}
    w_{4}^{(x^{i})} (t+1) = w_{4}^{(x^{i})} (t) - \eta \|x - x''\|_{2}^{2}
\label{equation_w4}
\end{equation}

As previously discussed, the replication of later-acquired sounds relies on the accumulation of various sound elements obtained during a crucial developmental period in early life, achieved through the incorporation of multiple acquired sounds, which we term "constituents". In the subsequent experiment, we set the number of constituents $N$ to 10. It's noteworthy that performance tends to enhance with a greater number of constituents ($N \leq |Z|$).

\begin{equation}
\begin{gathered}
X' = \delta_{i_n}(Z) \cdot W_{3} \\ 
\text{where} \quad \delta_{i_n}(Z) = \begin{cases}
    1, & \text{for the indices of N highest values in Z}  \\
    0, & \text{otherwise}
\end{cases} 
\end{gathered}
\end{equation}

\begin{equation}
X'' = W_{4} \cdot X'
\end{equation}

\section{Experiments}
In this section, we will detail the learning process of the models, along with the design and results of experiments. 

\subsection{Datasets}
As previously mentioned, our objective is to investigate the impact of the sound learning timing on second language sound acquisition. Accordingly, we utilize three distinct language datasets: English (sourced from \cite{Panayotov2015}), French (recorded by French speakers, encompassing both male and female voices), and Chinese (sourced from \cite{Magicdata2019}). Here, English serves primarily as the training dataset, thus regarded as L1, while the other two languages serve as L2.

\subsection{Training details}
The English training dataset is a 16 minutes' 16kHz multi-person read English speech sounds, randomly chosen among the original 1000 hours corpus of read English speech \cite{Panayotov2015}. They are transformed into more than 41000 20-dimensional MFCC using $Librosa$ Python library by setting $hop\_length=512$ (number of samples between successive frames), and $n\_fft=1024$ (length of the FFT window). These MFCCs are then fed into layer Y (SOM). We trained the layer Y for 100000 iterations. It is noticeable that the radius of the neighborhood of BMU changed along time. Specifically, here we apply Mexican hat function as the neighborhood function. 

As for the generative model, we initially configure the size of layer $Z$ to be 5000, while setting the threshold value at 0.00002 for adapting to an empty state or adding a new neuron, which is subsequently identified as the winner neuron $i_0$. This mechanism is based on the hypothesis that the variability of activity within a Self-Organizing Map (SOM) can serve as an indicator to discern whether the observed sound is novel or previously learned. In this hypothesis, when encountering a new sound, the activity pattern within the SOM would differ significantly, potentially necessitating the involvement of different neurons in its reconstruction compared to a familiar sound. Essentially, the SOM's ability to adapt its neural activations in response to new stimuli allows it to distinguish between learned and novel sounds based on the variance of this activity. It is crucial to acknowledge that increasing the threshold value enhances the dynamism of layer Z, but simultaneously, it exacerbates the problem of poor generalization.

\subsection{Results}
In this section, we will represent and analyse experimental results obtained from the training and the designed experiments talked previously.

\subsubsection{Self-organised map}
\begin{figure}[h!]
  \centering
  \begin{subfigure}[b]{0.48\textwidth}
    \includegraphics[width=\textwidth]{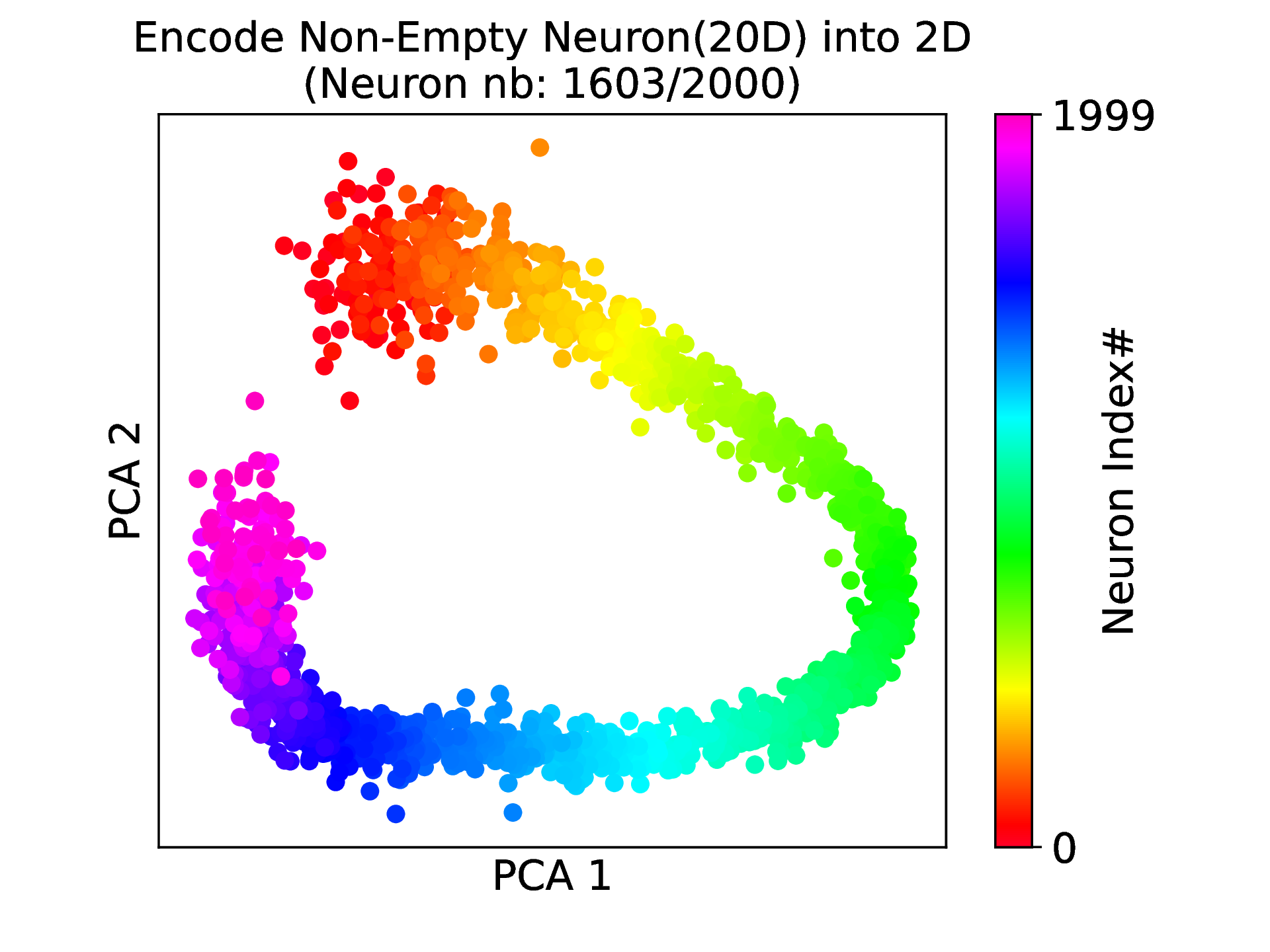}
    \caption{Visualizing Dimension Reduction of SOM Neurons through PCA} 
    \label{neuron_PCA}
  \end{subfigure}
  \hfill
  \begin{subfigure}[b]{0.48\textwidth}
    \includegraphics[width=\textwidth]{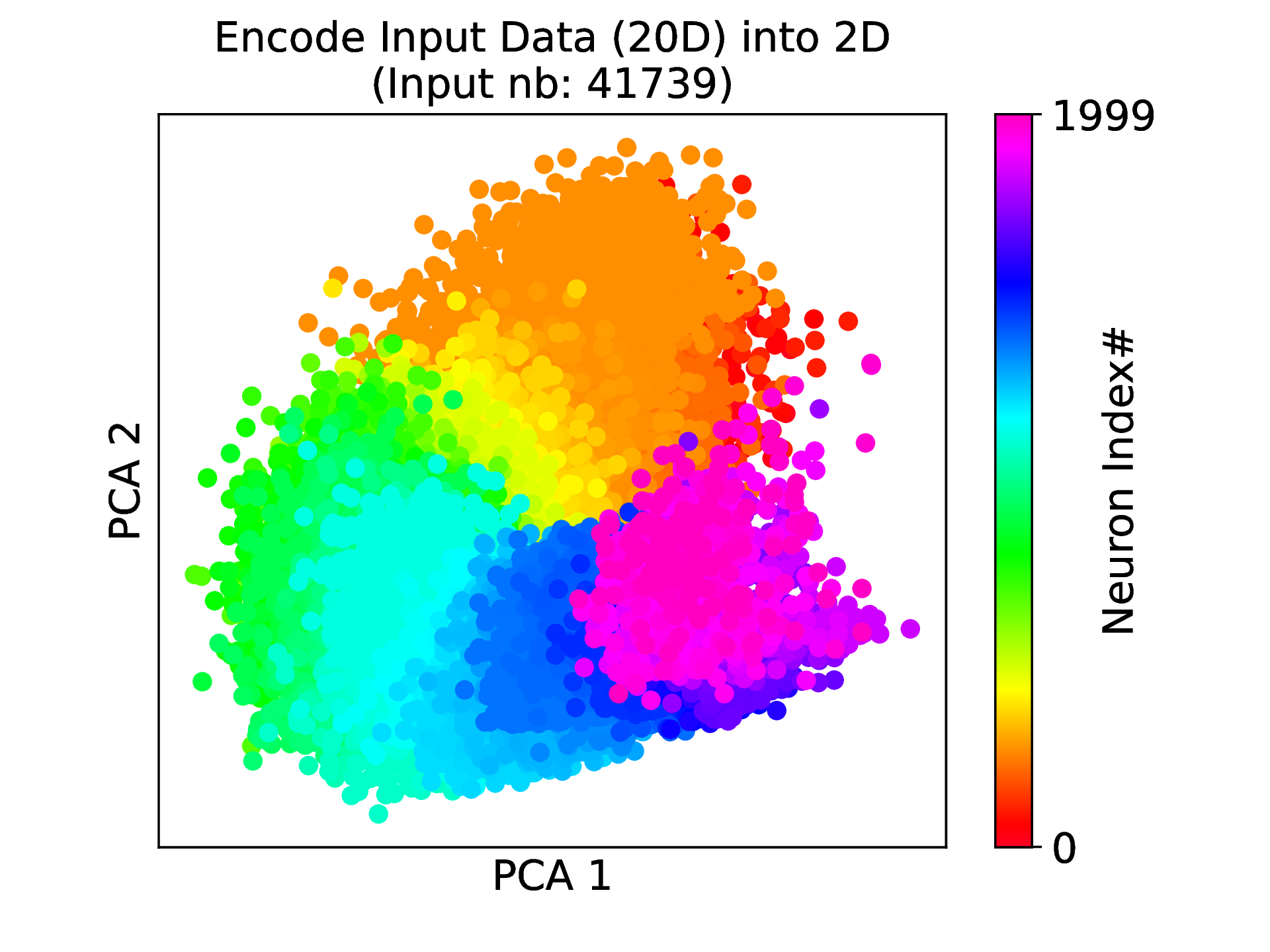}
    \caption{Visualizing Dimension Reduction of MFCC inputs through PCA} 
    \label{input_pca}
  \end{subfigure}
  \caption{Self-organised map training results}
  \label{som_results}
\end{figure}

A fundamental characteristic of SOM is that neurons with geographic proximity encode similar inputs. As such,
in Fig. \ref{neuron_PCA}, high-dimensional neurons are projected into a 2-dimensional space using PCA (Principal Components Analysis) for visualization, with each dot representing a neuron. It's noteworthy that only neurons coding for at least one input X are depicted (1603 out of 2000). Dots are displayed in various colors, and colors that are closely aligned according to the colorbar next to the figure signify geographic proximity. Similarly, Fig. \ref{input_pca} encodes inputs into a 2-dimensional space using PCA. However this time, each dot represents one input MFCC data (totaling more than 41000), with the color of each dot corresponding to the color of the winning neuron. Thus, inputs sharing the same neuron are depicted in the same color. Additionally, inputs whose winner neurons are neighbors, though not shared, should exhibit similar colors based on the colorbar next to the figure.
The analysis of Fig. \ref{som_results} yields two notable observations. Firstly, in Fig. \ref{neuron_PCA}, closely located dots exhibit similar colors as indicated by the colorbar, suggesting that neighboring neurons are indeed close to each other in the original 20-dimensional space. Additionally, the color variations of the dots align with those of the colorbar, reinforcing the proximity of neighboring neurons. Secondly, in Fig. \ref{input_pca}, inputs belonging to the same clustering are encoded by the same neuron or neighboring neurons, evident from the shared or similar colors among the dots. This correspondence between input clusters and neuron encoding further validates the effectiveness of the SOM learning process. Therefore, we can confidently conclude that the SOM has learned effectively 
to discriminate the input MFCC.

\subsubsection{Pattern similarity and sound reconstruction}
\begin{figure}
  \centering
    \begin{subfigure}[b]{0.32\textwidth}
    \includegraphics[width=\textwidth]{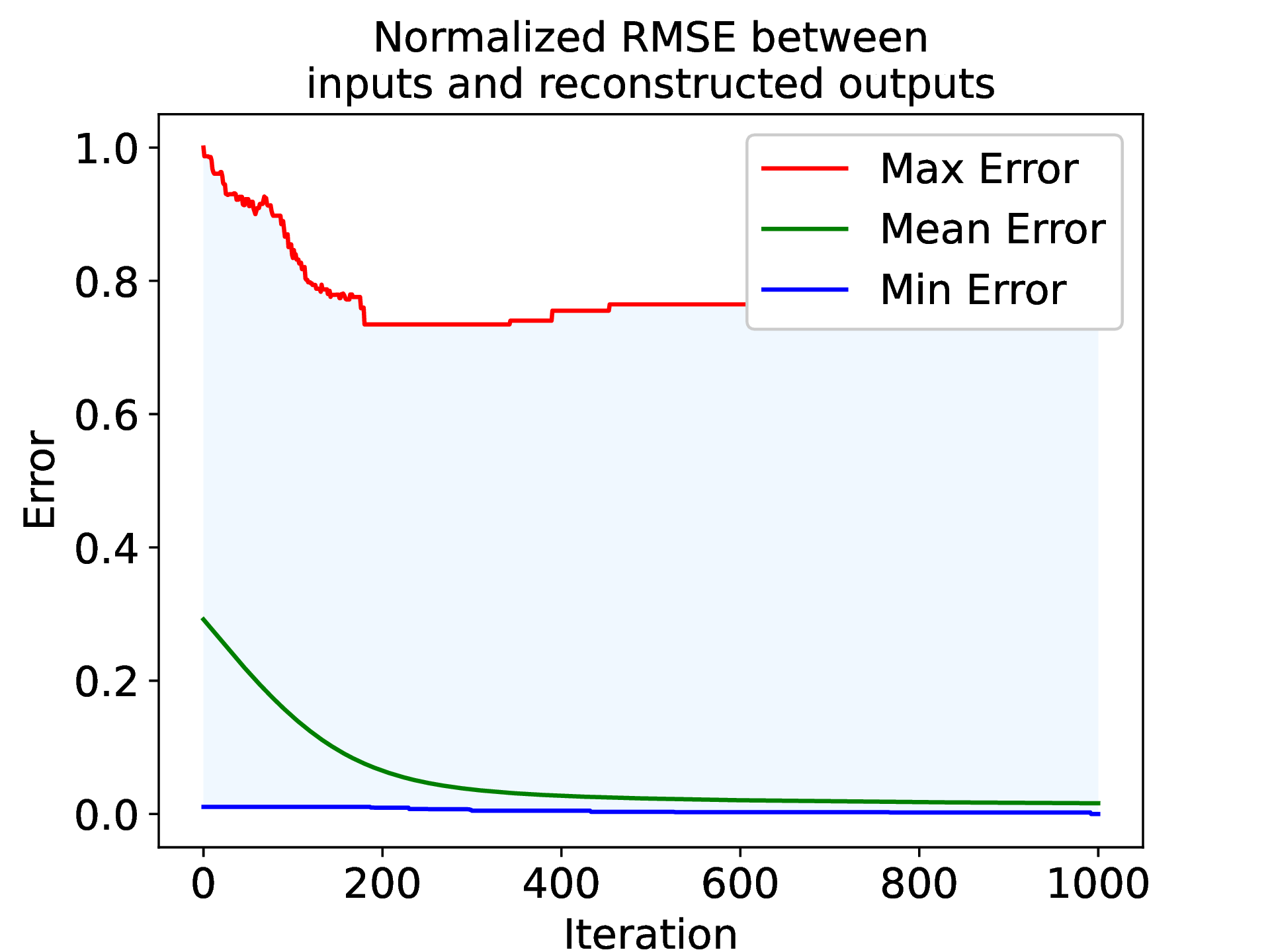}
    \caption{Training error} 
    \label{recon_error_train}
    \end{subfigure}
    \begin{subfigure}[b]{0.32\textwidth}
    \includegraphics[width=\textwidth]{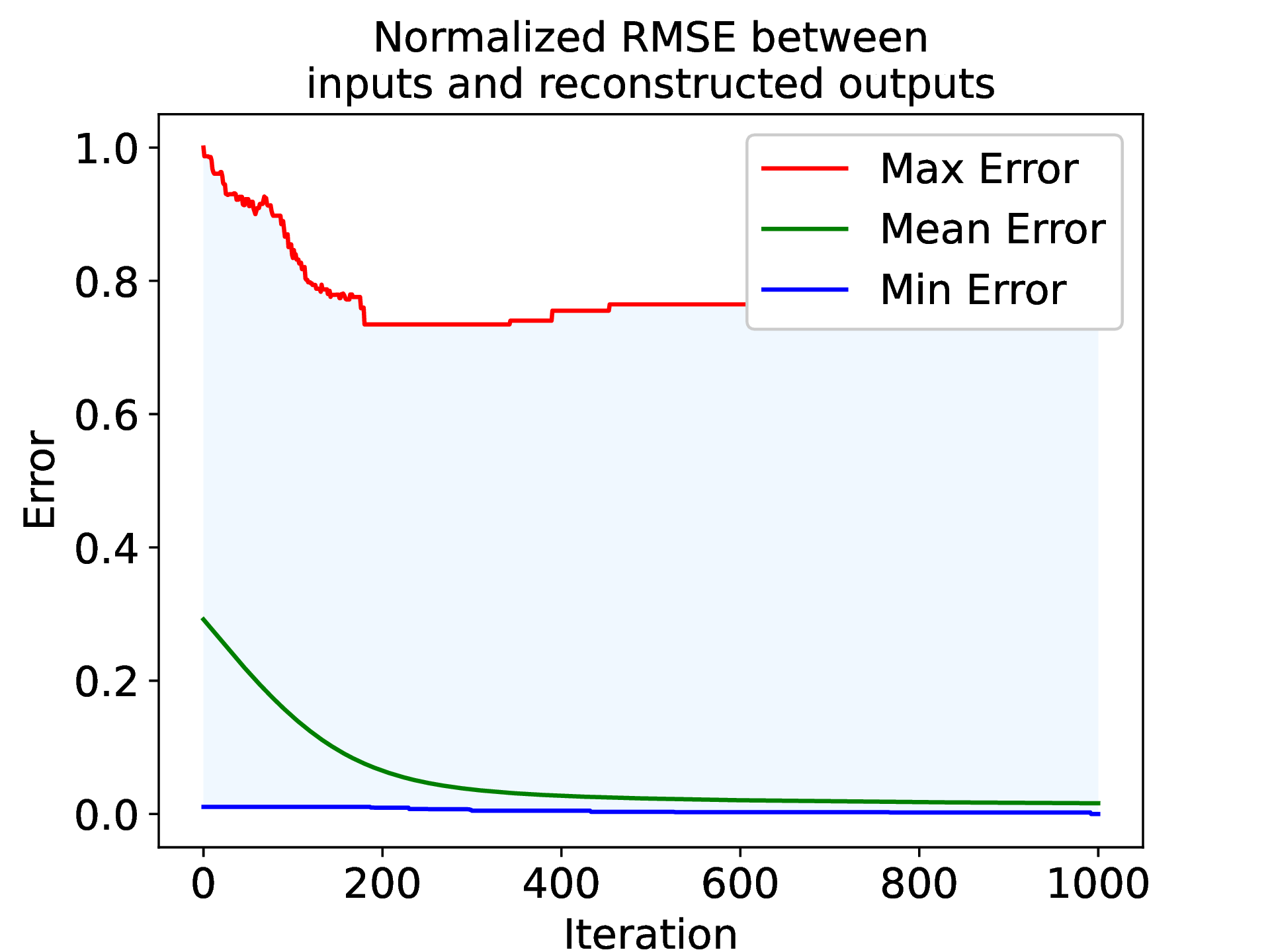}
    \caption{Testing error} \label{recon_error_test}
    \end{subfigure}
    \begin{subfigure}[b]{0.32\textwidth}
    \includegraphics[width=\textwidth]{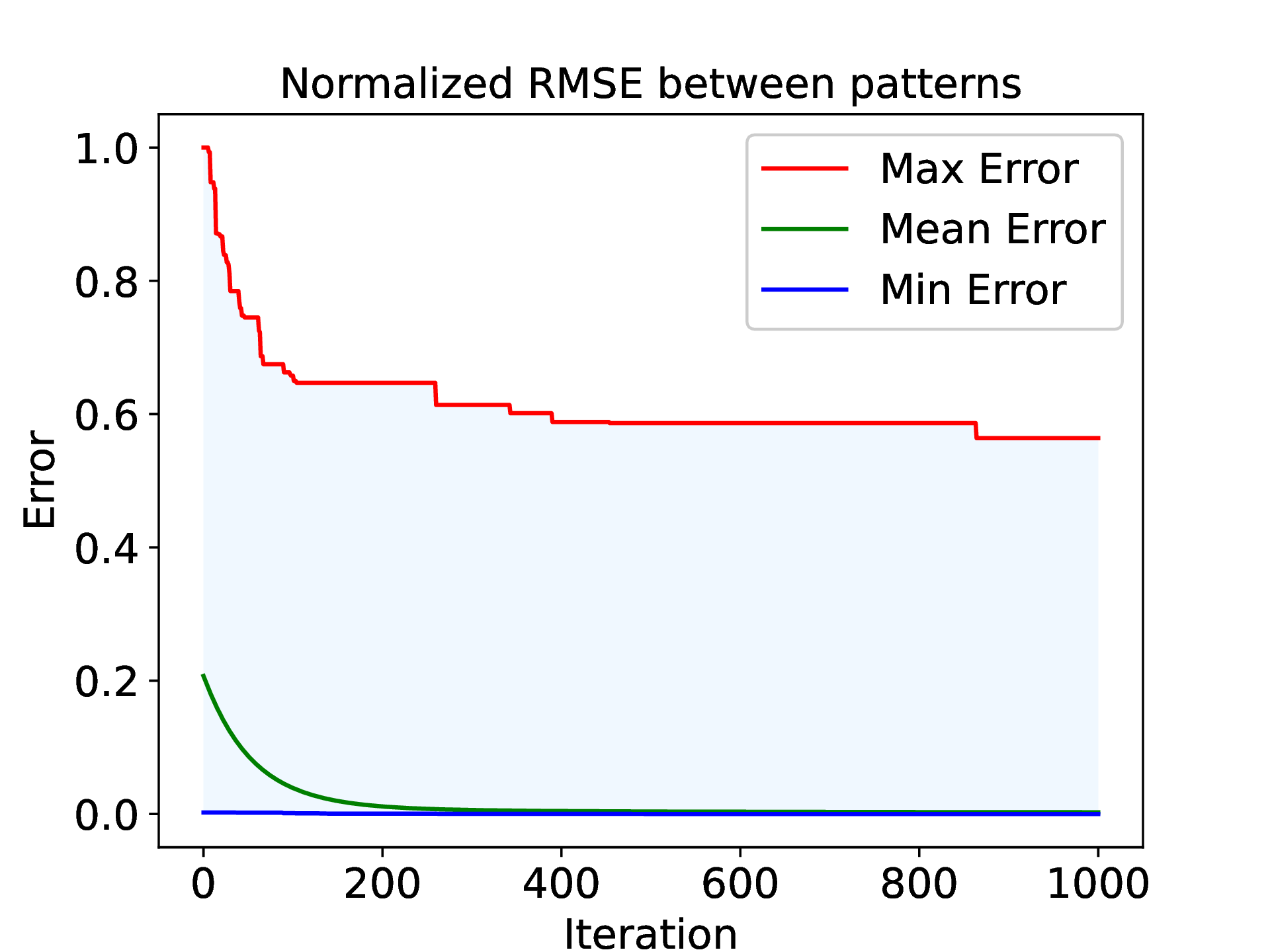}
    \caption{Training pattern error} \label{pattern_error_train}
    \end{subfigure}
  \caption{The trend of error between input MFCC and reconstructed MFCC for both training and test datasets (Fig. \ref{recon_error_train} and Fig. \ref{recon_error_test} respectively). The model employs a mechanism to adjust randomly generated Gaussian-type inputs to approximate the pattern of the reproduced sound to that of the heard sound, resulting in decreased error over time, as illustrated in Fig. \ref{pattern_error_train}.}
  \label{training_errpr}
\end{figure}

In the CL mode depicted in Fig. \ref{schema}, symbolizing the speech sound learning of newborns during early infancy, it is imperative to evaluate its ability to 
reproduce
input sounds. Therefore, it is essential to assess the efficiency and effectiveness of the model in reproducing MFCC inputs. Fig. \ref{recon_error_train} illustrates the trend of error between the input MFCC in the training dataset and the reconstructed MFCC. Here, we observe a decrease in error over time, with the mean error among 41000 inputs approaching 0. A similar trend is evident in the test dataset, as depicted in Fig. \ref{recon_error_test}. Additionally, as mentioned earlier, a randomly generated Gaussian-type input is added to itself if this input amplifies the activity of the winner neuron, as described in Equation \ref{equation_w3}. This mechanism aims at approximating the pattern of the reproduced sound to that of the heard sound. Consequently, if the error of reproduction decreases, so should the error between
these two patterns. This error is visualized in Fig. \ref{pattern_error_train}, indicating a decrease over time, eventually approximating 0 by the end of 1000 iterations.

\subsubsection{Mono/bilinguality and reconstruction mode comparison}

In addition to its ability to imitate sounds regardless of language, the model should excel at capturing the inherent differences between languages. This capability is crucial for exploring L1 or L2 language learning during different age of development. In these experiments, we investigate the differences between models that learn solely one language and those that learn two languages. 

As mentioned earlier, we propose two modes, CO and CL, representing sound imitation during different developmental stages: early infancy (before one year of age) versus later infancy or adulthood. Consequently, we have designed four conditions to independently investigate the effects of minimal input and critical period. Specifically, for the minimal input experiment, we compare the ability to imitate L2 sounds between models trained on both L1 and L2 and models trained solely on L1. For the critical time experiment, we further compare the imitation capability of L2 sounds separately under the CO and CL modes, which correspond to later infancy and early infancy respectively.

\begin{figure}[h!]
  \centering
  \begin{subfigure}[b]{0.45\textwidth}
    \includegraphics[width=\textwidth]{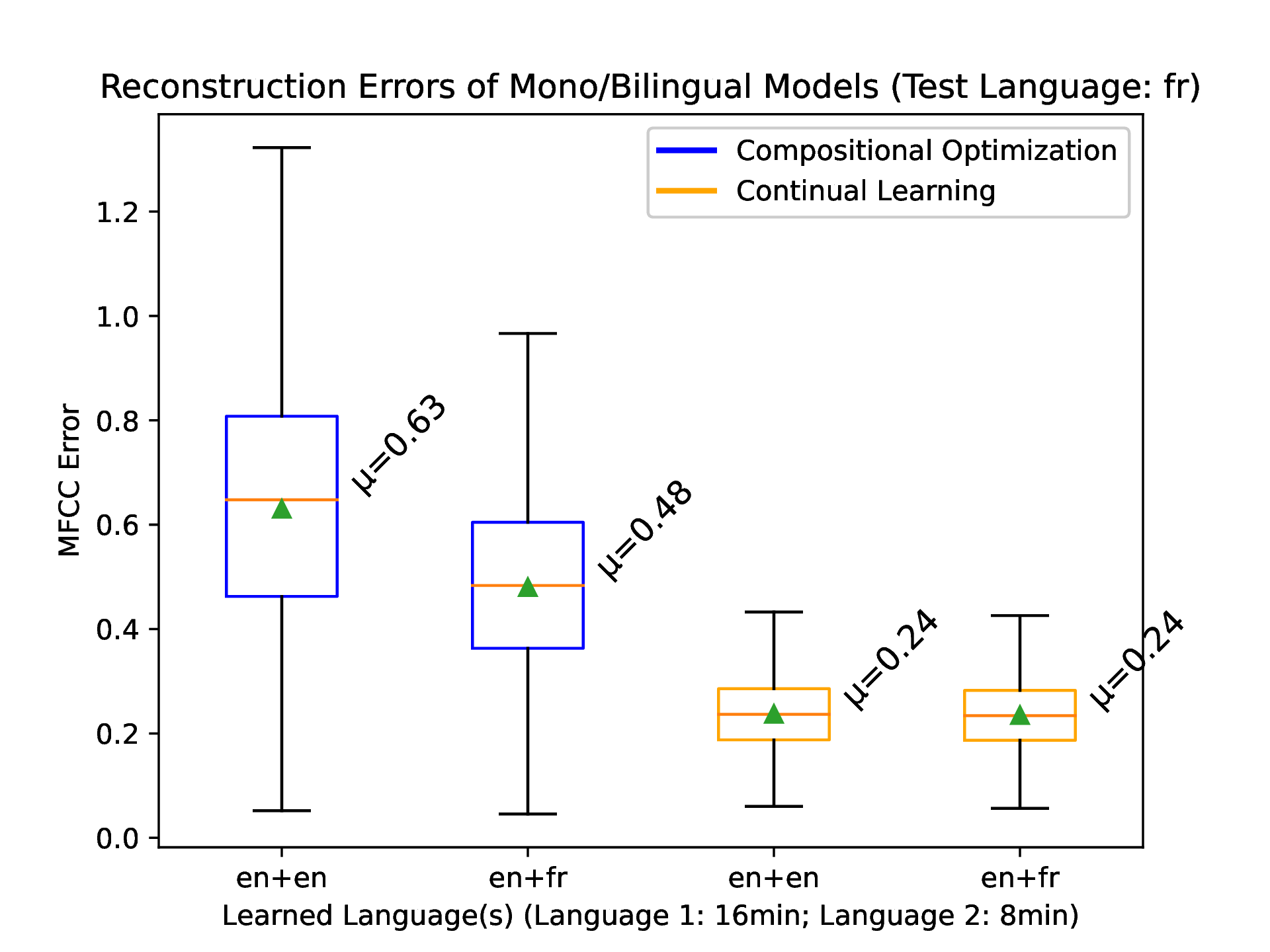}
    \caption{Model and mode comparison with French as the second language} 
    \label{modelComparaison_fr}
  \end{subfigure}
  \hfill  
  \begin{subfigure}[b]{0.45\textwidth}
    \includegraphics[width=\textwidth]{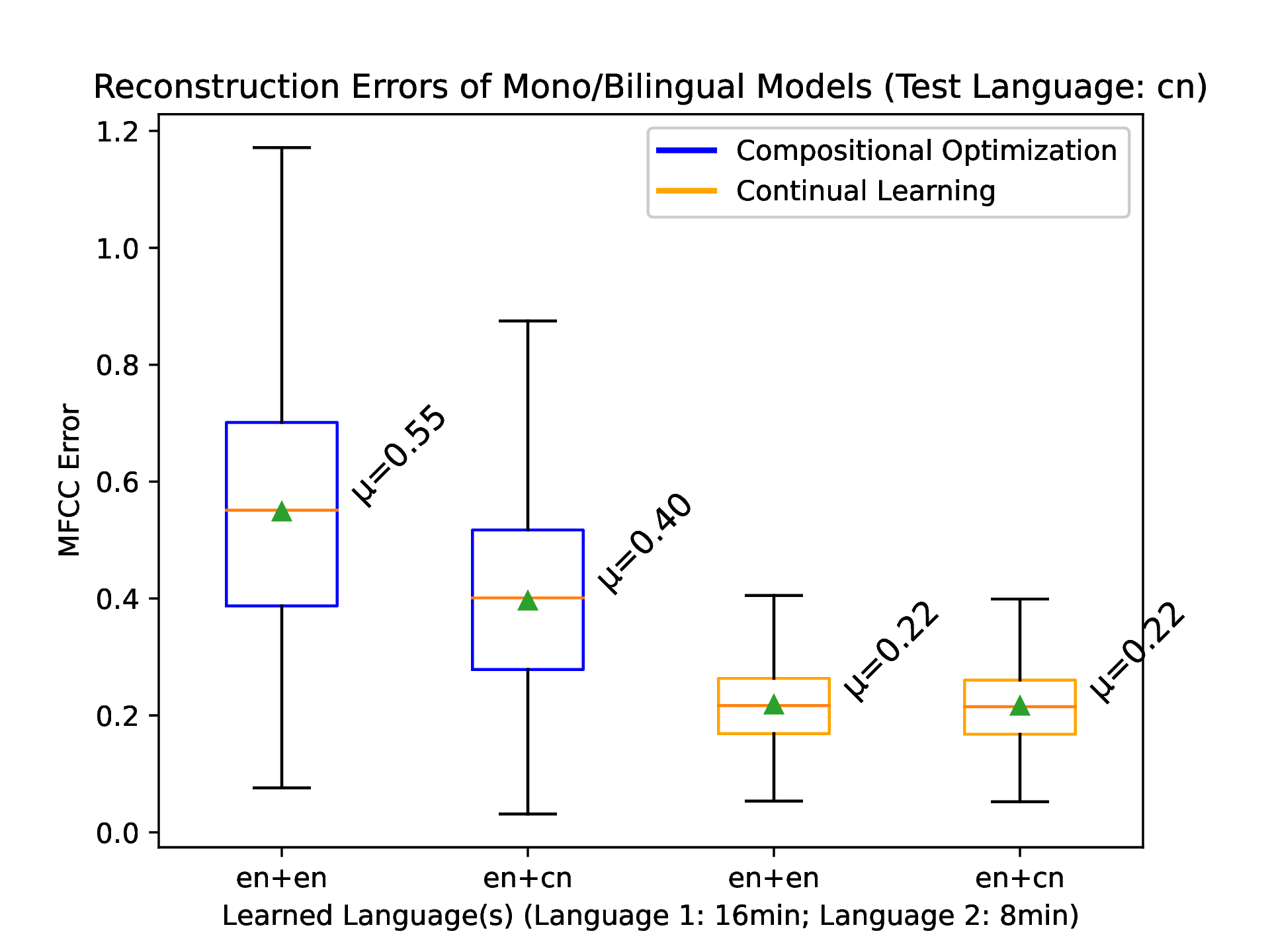}
    \caption{Model and mode comparison with Chinese as the second language} 
    \label{modelComparaison_cn}
  \end{subfigure}
  \caption{Critical period: Analysis of performance trends within the left bars of Fig. \ref{modelComparaison_cn} and Fig. \ref{modelComparaison_fr}, corresponding to models solely learning English, reveals that CL outperforms CO. This suggests that learning L2 after the critical period presents greater challenges in achieving comparable performance to learning during this critical period, aligning with the hypothesis of perceptual narrowing. Minimal input: Comparing results within the blue boxes, it's evident that errors decrease when L2 is learned, indicating that learning L2 further optimizes the model. A control group using L1 English as L2 was introduced (left box in each group with the horizontal axis labeled 'en+en'), where reconstruction error under CO mode disparities among L2 languages (Chinese $\approx$ French $>>$ English) indicate that the reduction in error is more likely due to dataset quality rather than quantity.}
  \label{modelComparaison}
\end{figure}

In the following experiment, we test the mode of reconstruction with models that learn English as L1 and French or Chinese as L2. As we can see in Fig. \ref{modelComparaison}, there are 2 modes of reconstruction, and on the left side of each mode is the model that has learned only English whereas on the right side has learned 8min L2 French or Chinese in addition to the 16min English. The blue boxes reconstruct MFCC outputs based on compositional optimization (constituents number $N=10$) whereas the orange ones based on continual learning. Please note that the test input language in this experiment is L2.

Focusing on the left bars within each group in Fig. \ref{modelComparaison_cn} and Fig. \ref{modelComparaison_fr}, which corresponds to the model that learned English solely, we observe the following performance trend across modes of reconstruction: CL $>$ CO. This can be interpreted as follows: when the acquisition of L2 occurs after the critical period (CO mode), it is more challenging to achieve performance comparable to learning during this period (CL mode). This result aligns with the perceptual narrowing phenomenon mentioned earlier.

If we compare the results within the group of blue boxes, we notice that errors of reconstruction decrease when L2 is learned, while errors in CL groups remain the same, as L2 has already been learned in both cases in the orange box group (L2 continual learning on L1 learned model (left orange box) is the same as L1+L2 learned model (right orange box)). It is worth noting that even after learning L2 in the CO mode, the error is still higher than in the CL mode (cn: $0.38 > 0.21$; en: $0.47 > 0.22$). This is due to the number of constituents not being optimized. However, the main focus here is on the decrease between the two blue boxes, where the minimal input of them is L1 only (left blue box) versus L1+L2 (right blue box), thus further supporting the perceptual narrowing hypothesis.

One may question whether the reduction in error within the CO group in Fig. \ref{modelComparaison_cn} and Fig. \ref{modelComparaison_fr} is attributed to the increased size of the learning dataset rather than linguistic features alone. To address this, we introduced a control group by considering L1 English as the L2, as depicted on the left of each group with the horizontal axis labeled 'en+en'. The disparities in reconstruction error under CO mode among the L2 languages are as follows: Chinese $\approx$ French $>>$ English. This observation suggests that the reduction in error is more likely due to the quality rather than the quantity of the dataset.
\begin{figure}[h!]
\centering
\includegraphics[width=0.8\textwidth]{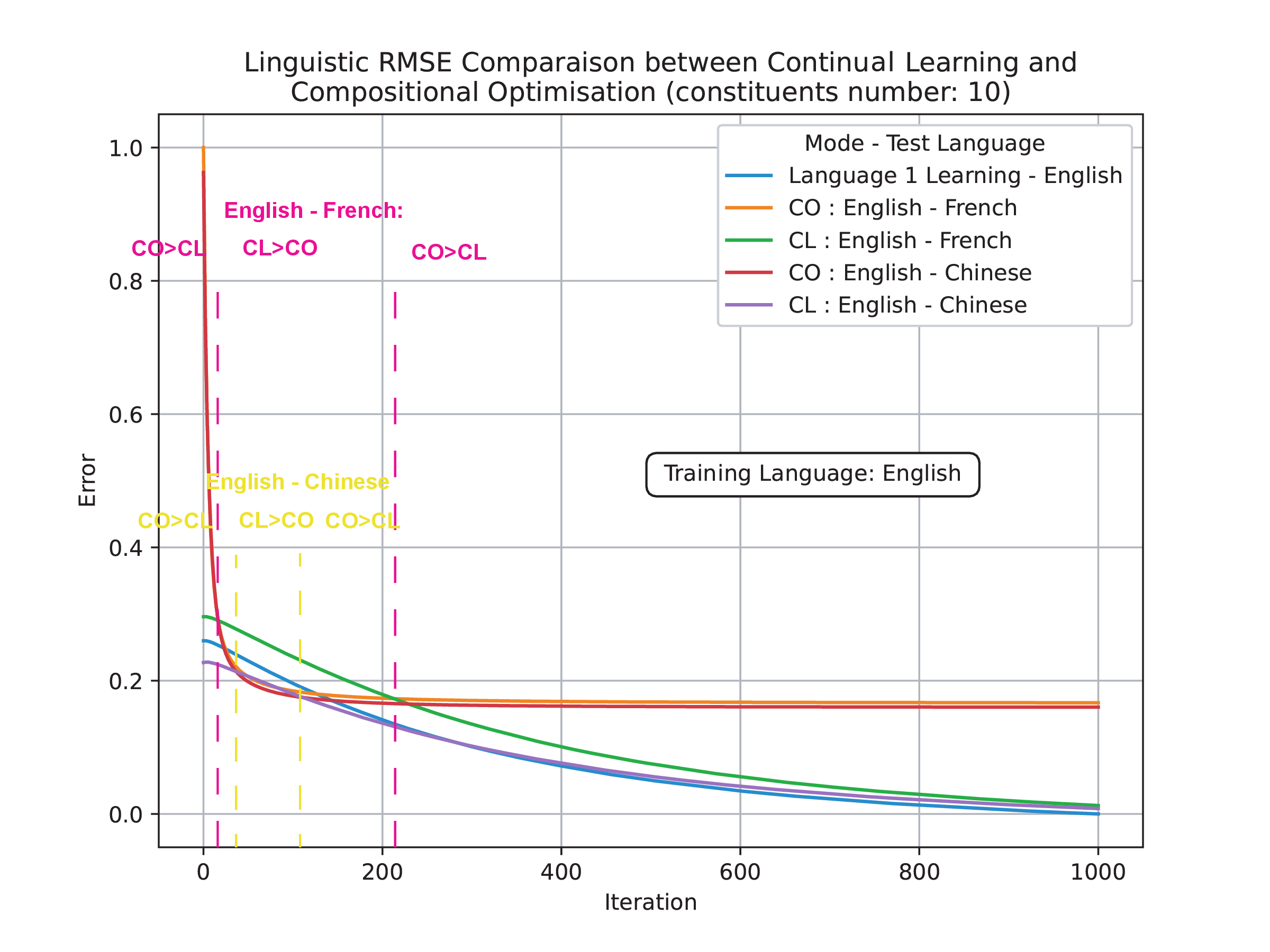}
\caption{Model Reconstruction Comparison: CO vs CL. Both models start from a baseline model that learned the same 16-minute training English Dataset. By the end of 1000 iterations, the reconstruction error of CL approximates to 0, which is significantly lower than the CO mode that underwent no learning.} 
\label{linguisticOptimization}
\end{figure}

Consistently, in Fig. \ref{linguisticOptimization}, we depict the trend of errors between CO and CL modes while reconstructing L2 language input sound. As observed, both modes of reconstruction exhibit a significant decline over time. However, by the end of 1000 iterations, the error of the CL mode becomes much smaller than that of the CO mode. This result is consistent with the findings we presented earlier.

\subsubsection{Catastrophic forgetting}

\begin{figure}[h!]
\centering
\includegraphics[width=0.8\textwidth]{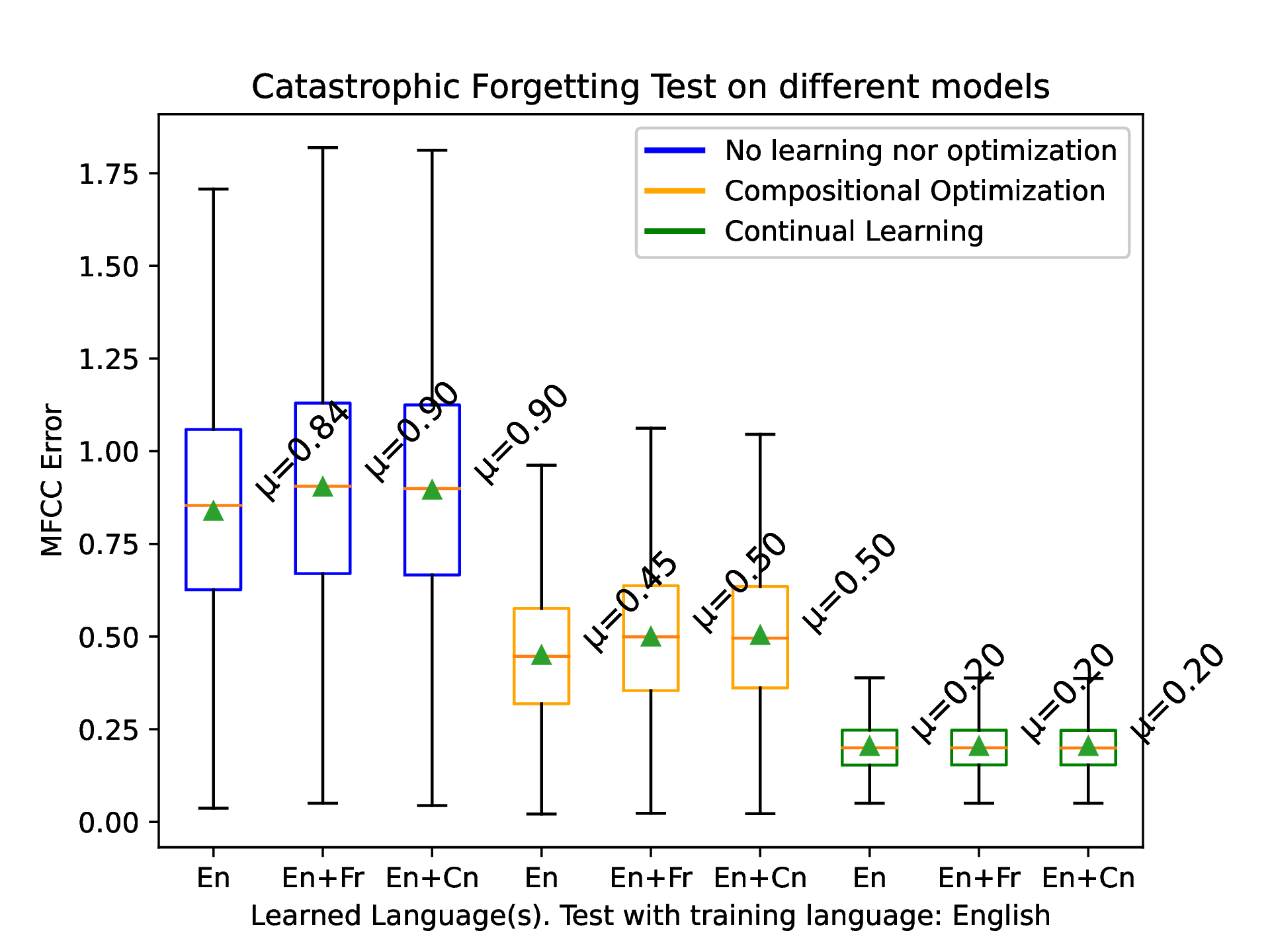}
\caption{We conducted a catastrophic forgetting test on three different models, each tested under different conditions: monolingual (L1: English learned only) or bilingual (L1: English; L2: French or Chinese). The test dataset is in L1. Subtle differences were observed within each color group across all reconstruction modes, suggesting that our model may not be prone to catastrophic forgetting.} 
\label{catastrophicForgetting}
\end{figure}

Catastrophic forgetting poses a significant challenge in neural networks and algorithms \cite{Goodfellow2014}. It refers to the phenomenon where a model tends to erase previously acquired knowledge when learning new information. In this experiment, we aim to determine if our proposed neural network is susceptible to this issue. We compare the L1 reconstruction capacity between models trained solely on L1 (with X axis value 'En') and those trained on first L1 and then on L2 (with X axis value 'En+Fr' or 'En+Cn'), irrespective of reconstruction mode. Theoretically, if the proposed neural networks suffer from catastrophic forgetting, L1 reconstruction errors of models that learned L1 for 16 minutes should be much lower than those of models that first learned L1 for 16 minutes and then 8 minutes of L2, regardless of the mode group. However, as depicted in Fig. \ref{catastrophicForgetting}, only subtle differences are observed across all reconstruction modes, suggesting that our model may not suffer from catastrophic forgetting.

One potential explanation for the resilience of our neural network against catastrophic forgetting could be attributed to the mechanism where empty (or new) neurons are activated (or added) based on the variance of neuron activity in the self-organizing map.

\section{Discussion}
In this paper, we introduced a simple neural network trained on a limited dataset comprising 16 minutes of L1 data and 8 minutes of L2 data. Our primary objective was to investigate two hypotheses. Firstly, we sought to determine whether early infancy imitation of speech sounds contributes to later foreign language development. To explore this, we designed an encoder-decoder neural network capable of reconstructing speech sound inputs represented as MFCC. In addition to being able to reconstruct sound fed from input, to avoid the possibility that such performance was due to overfitting, we introduced a mono-bilingual model and compared its performance. Our results showed that, on one hand, models that learned L2 additionally based on the L1 baseline model outperformed L1 models in L2 performance. We introduced a control group where L1 played the L2 role to eliminate the possibility that such better performance was due to the increased size of the learning dataset. On the other hand, we also showed through experiments that having learned L2 did not lead to catastrophic forgetting in L1. These results indicate that our neural networks excel at capturing the inherent differences between languages to perform imitation. Secondly, we investigated whether the concepts of minimal input and critical period, closely associated with the perceptual narrowing phenomenon, influence infancy sound imitation. We designed two modes of reconstruction to simulate learning in different periods: Continual learning mode for language sound acquisition during the critical time window versus compositional optimization mode for language sound acquisition after the critical period. Our experiments were able to reveal that if second language acquisition occurs during later childhood or adulthood, learning foreign languages becomes more challenging.

As discussed previously, the neural networks we proposed are small-sized neural networks, where both the number of neurons and layers are much smaller than deep neural networks, which typically require substantial offline training periods. Differently, our proposed predictive coding-based continuous learning mode is a form of online learning, and its ability to incorporate new information in real-time without necessitating a complete retraining of the model ensures that online learning systems remain relevant and accurate even as the data evolves, which can be challenging for pre-trained deep networks to handle effectively.

Indeed, one may notice that we did not compare the imitation performance between deep networks and our proposed neural networks. However, the objective of our study is to develop a simple, explanatory model rather than to achieve the highest performance metrics. Our focus is on creating a model that is sufficiently straightforward to elucidate the underlying infant sound learning mechanism. Unlike complex deep learning models that prioritize performance at the expense of transparency, our approach aims to strike a balance between simplicity and functionality. By doing so, we enable a clearer understanding of how the model processes and interprets data, which is crucial for applications where interpretability and insight into the learning process are as important as the outcomes themselves. This emphasis on simplicity and explanatory power ensures that the model is not only effective but also accessible and informative.

In the future, we intend to further develop this neural network for language sound sequence (phonotactic patterns) learning, focusing on addressing the varying phoneme-sets between L1 and L2 languages, as well as the complexity introduced by tonal variations, especially in languages like Chinese.

\section{Acknowledgement}
Xiaodan Chen is supported by CYU-IPAL and the A*STAR Research Attachment Programme. Mathias Quoy was supported by CNRS funding for a research semester at the IPAL Lab in Singapore. We also acknowledge the use of the Osaka compute-cluster resources of CY Cergy Paris University.

%
%
%

\bibliography{Bibliography}
\bibliographystyle{splncs04}

\end{document}